\documentclass[11pt,a4paper]{article}
\usepackage[hyperref]{acl2018}
\usepackage{times}
\usepackage{latexsym}
\usepackage{multirow}
\usepackage{graphicx}
\usepackage[utf8]{inputenc}

\usepackage{url}

\aclfinalcopy 
\def\aclpaperid{437} 

\newcommand{\items}{\ensuremath{{\mathcal I}}}
\newcommand{\allowed}{\ensuremath{{\mathcal A}}}
\newcommand{\infer}[2]{\ensuremath{\frac{#1}{#2}}}
\newcommand{\bconst}{\ensuremath{\overline{B}}}
\newcommand{\econst}{\ensuremath{\overline{E}}}

\newcommand{\softmax}{\ensuremath{\mathrm{softmax}}}
\newcommand{\NULL}{\ensuremath{\mathrm{NULL}}}
\newcommand{\w}{\textbf{w}}

\newcommand{\TAG}[1]{\ensuremath{{\mathcal #1}}}

\usepackage{microtype}

\usepackage{array,multirow}

\usepackage{arydshln}

\title{Generalized chart constraints for efficient PCFG and TAG parsing}

\author{
Stefan Grünewald \and Sophie Henning \and Alexander Koller\\
Department of Language Science and Technology\\
Saarland University, Saarbrücken, Germany\\
\url{{stefang|shenning|koller}@coli.uni-saarland.de}
}

\date{}

\begin{document}
\maketitle

\begin{abstract}
  Chart constraints, which specify at which string positions a
  constituent may begin or end, have been shown to speed up chart
  parsers for PCFGs. We generalize chart constraints to more
  expressive grammar formalisms and describe a neural tagger which
  predicts chart constraints at very high precision. Our constraints
  accelerate both PCFG and TAG parsing, and combine effectively with
  other pruning techniques (coarse-to-fine and supertagging) for an
  overall speedup of two orders of magnitude, while improving
  accuracy.
\end{abstract}


\section{Introduction} \label{sec:introduction}

Effective and high-precision pruning is essential for making
statistical parsers fast and accurate. Existing pruning techniques
differ in the source of parsing complexity they tackle. Beam search
\cite{Collins03} bounds the number of entries in each cell of the
parse chart; supertagging
\cite{Bangalore:1999:SAA:973306.973310,clark07:_wide_cover_effic_statis_parsin,lewis16:_lstm_ccg_parsin}
bounds the number of lexicon entries for each input token; and
coarse-to-fine parsing \cite{CharniakCF06} blocks chart cells that
were not useful when parsing with a coarser-grained grammar.

One very direct method for limiting the chart cells the parser
considers is through \emph{chart constraints}
\cite{roark12:_finit_state_chart_const_reduc}: a tagger first
identifies string positions at which constituents may begin or end,
and the chart parser may then only fill cells which respect these
constraints. Roark et al.\ found that begin and end chart constraints
accelerated PCFG parsing by up to 8x. However, in their original form,
chart constraints are limited to PCFGs and cannot be directly applied
to more expressive formalisms, such as tree-adjoining grammar (TAG,
\newcite{joshi;etal1997}).

Chart constraints prune the ways in which smaller structures can be
combined into bigger ones. Intuitively, they are complementary to
supertagging, which constrains lexical ambiguity in lexicalized
grammar formalisms such as TAG and CCG, and has been shown to
drastically improve efficiency and accuracy for these
\cite{Bangalore:2009:MPD:1620853.1620904,lewis16:_lstm_ccg_parsin,kasai17:_tag_parsin_neural_networ_vector_repres_super}. For
CCG specifically,
\newcite{zhang10:_chart_prunin_fast_lexic_gramm_parsin} showed that
supertagging combines favorably with chart constraints. To our
knowledge, similar results for other grammar formalisms are not
available.



In this paper, we make two contributions. First, we generalize chart
constraints to more expressive grammar formalisms by casting them in
terms of \emph{allowable parse items} that should be considered by the
parser. The Roark chart constraints are the special case for PCFGs and
CKY; our view applies to any grammar formalism for which a parser can
be specified in terms of parsing schemata.  Second, we present a
neural tagger which predicts begin and end constraints with an
accuracy around 98\%. We show that these chart constraints speed up a
PCFG parser by 18x and a TAG chart parser by 4x. Furthermore, chart
constraints can be combined effectively with coarse-to-fine parsing
for PCFGs (for an overall speedup of 70x) and supertagging for TAG
(overall speedup of 124x), all while improving the accuracy over those
of the baseline parsers. Our code is part of the Alto
parser \cite{alto-demo-17}, available at \url{http://bitbucket.org/tclup/alto}.


\section{Generalized chart constraints} \label{sec:pruning}

Roark et al.\ 
define \emph{begin}
and \emph{end} chart constraints. A begin constraint $\bconst$ for the
string \w\ is a set of positions in \w\ at which no constituent of
width two or more may start. Conversely, an end constraint $\econst$
describes where constituents may not end.

Roark et al.\ focus on speeding up the standard CKY parser for PCFGs
with chart constraints. They do this by declaring a cell $[i,k]$ of
the CKY parse chart as \emph{closed} if $i \in \bconst$ or
$k \in \econst$, and modifying the CKY algorithm such that no
nonterminals may be entered into closed cells. They show this to be
very effective for PCFG parsing; but in its reliance on CKY chart
cells, their algorithm is not directly applicable to other parsing
algorithms or grammar formalisms.

\subsection{Allowable items}

In this paper, we take a more general perspective on chart
constraints, which we express in terms of \emph{parsing schemata}
\cite{shieber95:_princ_implem_deduc_parsin}. A parsing schema consists
of a set \items\ of \emph{items}, which are derived from initial items
by applying inference rules. Once all derivable items have been
calculated, we can calculate the best parse tree by following the
derivations of the \emph{goal items} backwards.

Many parsing algorithms can be expressed in terms of parsing
schemata. For instance, the CKY algorithm for CFGs uses items of the
form $[A, i, k]$ to express that the substring from $i$ to $k$ can be
derived from the nonterminal $A$, and derives new items out of old
ones using the inference rule
$$\infer{[B, i, j] \quad [C, j, k] \quad A \rightarrow B\;
  C}{[A,i,k]}$$

The purpose of a chart constraint is to describe a set of
\emph{allowable items} $\allowed \subseteq \items$. We restrict the
parsing algorithm so that the consequent item of an inference rule may
only be derived if it is allowable. If all items that are required for
the best derivation are allowable, the parser remains complete, but
may become faster because fewer items are derived.

For the specific case of the CKY algorithm for PCFGs, we can simulate
the behavior of Roark et al.'s algorithm by defining an item $[A,i,k]$
as allowable if $i \not\in \bconst$ and $k \not\in \econst$.

\begin{figure}
  \centering
  \includegraphics[width=\columnwidth]{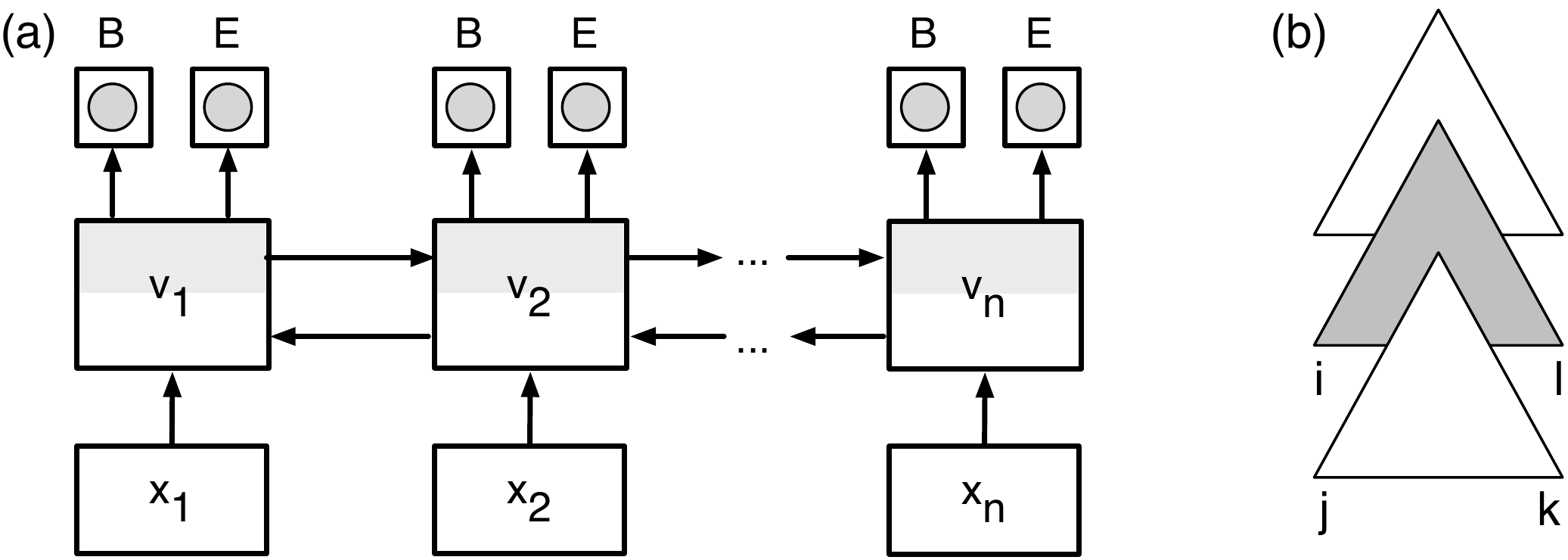}   \strut\\[-7ex]\strut
  \caption{(a) Chart-constraint tagger; (b) TAG adjunction.}
  \label{fig:adjunction}
\end{figure}

\subsection{Chart constraints and  binarization} 
\label{sec:binarization}

One technical challenge regarding chart constraints arises in the
context of binarization. Chart constraints are trained to identify
constituent boundaries in the original treebank, where nodes may
have more than two children. However, an efficient chart parser for
PCFG can combine only two adjacent constituents in each step. Thus, if
the original tree used the rule $A \rightarrow B\; C\; D$, the parser
needs to first combine $B$ with $C$, say into the substring $[i,k]$,
and then the result with $D$ (or vice versa). This intermediate
parsing item for $[i,k]$ must be allowable, even if $k \in \econst$,
because it does not represent a real constituent; it is only a
computation step on the way towards one.

We solve this problem by keeping track in the parse
items whether they were an intermediate result caused by binarization,
or a complete constituent.  This generalizes Roark et al.'s cells that
are ``closed to complete constituents''. For instance, when converting
a PCFG grammar to Chomsky normal form, one can distinguish the ``new''
nonterminals generated by the CNF conversion from those that were
already present in the original grammar. We can then let an item
$[A,i,k]$ be allowable if $i \not\in \bconst$ and either $k \not\in
\econst$ or $A$ is new.


\subsection{Allowable items for TAG parsing} \label{sec:pruning-tag}

By interpreting chart constraints in terms of allowable parse items,
we can apply them to a wide range of grammar formalisms beyond
PCFGs. We illustrate this by defining allowable parse items for
TAG. Parse items for TAG
\cite{shieber95:_princ_implem_deduc_parsin,kallmeyer10:_parsin_beyon_contex_free_gramm}
are of the form $[\TAG{X}, i, j, k, l]$, where $i,l$ are string
positions, and $j,k$ are either both string positions or both are
\NULL. $\TAG{X}$ is a complex representation of a position in an
elementary tree, which we do not go into here; see the literature for
details. The item describes a derivation of the string from position
$i$ to $l$. If $j$ and $k$ are \NULL, then the derivation starts with
an initial tree and covers the entire substring. Otherwise, it starts
with an auxiliary tree, and there is a gap in its string yield from
$j$ to $k$. Such an item will later be adjoined at a node which covers
the substring from $j$ to $k$ using the following inference rule (see
Fig.~\ref{fig:adjunction}b):
$$\infer{[\TAG{X}, i, j, k, l] \quad [\TAG{Y}, j, r, s, k]}{[\TAG{Y}', i, r, s, l]}$$

\noindent
Assuming begin and end constraints as above, we define allowable
TAG items as follows. First, an item $[\TAG{X},i,j,k,l]$ is not
allowable if $i \in \bconst$ or $l \in \econst$. Second, if $j$ and
$k$ are not \NULL, then the item is not allowable if $j \in \bconst$
or $k \in \econst$ (else there will  be no constituent from
$j$ to $k$ at which the item could be adjoined). Otherwise, the item
is allowable.

\subsection{Allowable states for IRTG  parsing} \label{sec:pruning-irtg}

Allowable items have a particularly direct interpretation when parsing
with Interpreted Regular Tree Grammars (IRTGs, \newcite{KollerK11}), a
grammar formalism which generalizes PCFG, TAG, and many others. Chart
parsers for IRTG describe substructures of the input object as
\emph{states} of a finite tree automaton $D$. When we encode a PCFG as
an IRTG, these states are of the form $[i,k]$; when we encode a TAG
grammar, they are of the form $[i,j,k,l]$. Thus chart constraints
describe \emph{allowable states} of this automaton, and we can prune
the chart simply by restricting $D$ to rules that use only allowable
states.

In the experiments below, we use the Alto IRTG parser
\cite{alto-demo-17}, modified to implement chart constraints as
allowable states. We convert the PCFG and TAG grammars into IRTG
grammars and use the parsing algorithms of \newcite{GroschwitzKJ16}:
``condensed intersection'' for PCFG parsing and the ``sibling-finder''
algorithm for TAG. Both of these implement the CKY algorithm and
compute charts which correspond to the parsing schemata sketched
above.

\section{Neural chart-constraint tagging}

Roark et al.\ predict the begin and end constraints for a string \w\
using a log-linear model with manually designed features. We replace
this with a neural tagger (Fig.~\ref{fig:adjunction}a), which reads
the input sentence token by token and jointly predicts for each string
position whether it is in \bconst\ and/or \econst. 

Technically, our
tagger is a two-layer bidirectional LSTM
\cite{kiperwasser16:_simpl_accur_depen_parsin_using,lewis16:_lstm_ccg_parsin,kummerfeld17:_parsin_traces}. In
each time step, it reads as input a pair $x_i = (w_i,p_i)$ of one-hot
encodings of a word $w_i$ and a POS tag $p_i$, and embeds them into
dense vectors (using pretrained GloVe word embeddings
\cite{pennington14:_glove} for $w_i$ and learned POS tag embeddings
for $p_i$). It then computes the probability that a constituent begins
(ends) at position $i$ from the concatenation
$v_i = v^{F2}_i \circ v^{B2}_i$ of the hidden states $v^{F2}$ and
$v^{B2}$ of the second forward and backward LSTM at position $i$:
$$\displaystyle\begin{array}{rcl}
  P(B \mid \w, i) &= &\softmax(W_B \cdot v_i + b_B)\\
  P(E \mid \w,i) &= &\softmax(W_E \cdot v_i + b_E)
\end{array}
$$
We let $\bconst = \{ i \mid P(B | \w, i) < 1 - \theta\}$; that is, the
network predicts a begin constraint if the probability of \bconst\
exceeds a threshold $\theta$ (analogously for \econst). The threshold
allows us to trade off precision against recall; this is important
because false positives can prevent the parser from discovering the
best tree.


\begin{figure}
  \centering
  {\small
  \begin{tabular}{l|lll|lll}
    &\multicolumn{3}{c|}{\bconst}&\multicolumn{3}{c}{\econst} \\
    $\theta$ & acc & prec & recall & acc & prec & recall  \\\hline
    0.5 & 97.6 & 97.4 & 97.8 & 98.1 & 98.7 & 98.7 \\
    0.9 & 96.7 & 98.8 & 95.2 & 97.2 & 99.4 & 96.7 \\
    0.99 & 93.7 & 99.6 & 87.9 & 93.0 & 99.7 & 90.5 \\
  \end{tabular}
  }
  \caption{Chart-constraint tagging accuracy.}
  \label{fig:tagging}
\end{figure}
\section{Evaluation} \label{sec:evaluation}

We evaluated the efficacy of chart-constraint pruning for PCFG and TAG
parsing. All runtimes are on an AMD Opteron 6380 CPU at 2.5 GHz, using
Oracle Java version 8. See the Supplementary Materials for
details on the setup.

\subsection{PCFG parsing}

We trained the chart-constraint tagger on WSJ Sections 02--21. The
tagging accuracy on WSJ Section 23 is shown in
Fig.~\ref{fig:tagging}. As expected, an increasing threshold $\theta$
increases precision and decreases recall. Precision and recall are
comparable to Roark et al.'s log-linear model for \econst. Our tagger
achieves 94\% recall for \bconst\ at a precision of 99\%, compared to
Roark et al.'s recall of just over 80\% -- without the feature
engineering effort required by their system.\footnote{Note that the
  numbers are not directly comparable because Roark et al.\ evaluate
  their tagger on Section 24.}

We extracted a PCFG grammar from a right-binarized version of WSJ
Sections 02--21 using maximum likelihood estimation, applying a
horizontal markovization of 2 and using POS tags as terminal symbols
to avoid sparse data issues. We parsed Section 23 using a baseline
parser which does not prune the chart, obtaining a low f-score of 71,
which is typical for such a simple PCFG. We also parsed Section 23
with parsers which utilize the chart constraints predicted by the
tagger (on the original sentences and gold POS tags) and the gold
chart constraints from Section 23. The results are shown in
Fig.~\ref{fig:pcfg-table}; ``time'' is the mean time to compute the
chart for each sentence, in milliseconds.

Chart constraints by themselves speed the parser up by factor of 18x
at $\theta=0.5$; higher values of $\theta$ did not increase the
parsing accuracy further, but yielded smaller speedups. This compares
to an 8x speedup in Roark et al.; the difference may be due to the
higher \bconst\ recall of our neural tagger. Furthermore, when we
combine chart constraints with the coarse-to-fine parser of
\newcite{irtg-ctf-17}, using their threshold of $10^{-5}$ for CTF
pruning, the two pruning methods amplify each other, yielding an
overall speedup of up to 70x.\footnote{Our CTF numbers differ slightly
  from Teichmann et al.'s because they only parse sentences with up to
  40 words and use a different binarization method.}

\setlength{\tabcolsep}{5pt}

\begin{figure}
  \centering
  {\small
  \begin{tabular}{l|rrrr}
    \hline
    Parser & f-score & time & speedup & \% gold\\\hline
    Unpruned & 71.0 & 2599 & 1.0x & 4.4\\ 
    CC ($\theta = {0.5}$) & 75.0 & 143 & 18.2x & 91.8 \\ 
    CC (gold) & 77.6 & 143 & 18.2x & 100.0 \\ 
    \hline 
    CTF & 67.6 & 194 & 13.4x & 20.1\\ 
    CTF + CC (${\theta}{=}0.5)$ & 72.4 & 37 & \textbf{70.1x} & 94.3 \\
    CTF + CC (gold) & 75.3 & 38 & 68.4x & 100.0 
                                           \\\hline
  \end{tabular}
}
  \caption{Results for PCFG parsing.}
  \label{fig:pcfg-table}
\end{figure}


\subsection{TAG parsing}

For the TAG experiments, we converted WSJ Sections 02--21 into a TAG
corpus using the method of
\newcite{chen04:_autom_tags_penn_treeb}. This method sometimes adjoins
multiple auxiliary trees to the same node. We removed all but the last
adjunction at each node to make the derivations compatible with
standard TAG, shortening the sentences by about 40\% on average. To
combat sparse data, we replaced all numbers by NUMBER and all words
that do not have a GloVe embedding by UNK.

The neural chart-constraint tagger, trained on the shortened corpus,
achieves a recall of 93\% for \bconst\ and 98\% for \econst\ at 99\%
precision on the (shortened) Section 00. We chose a value of
$\theta=0.95$ for the experiments, since in the case of TAG
parsing, false positive chart constraints frequently prevent the
parser from finding any parse at all, and thus lower values of
$\theta$ strongly degrade the f-scores.

We read a PTAG grammar \cite{resnik1992} with 4731 unlexicalized
elementary trees off of the training corpus, binarized it, and used it
to parse Section 00. This grammar struggles with unseen words, and
thus achieves a rather low f-score (see Fig.~\ref{fig:tag-table}).
Chart constraints by themselves speed the TAG parser up by 3.8x,
almost matching the performance of gold chart constraints. This
improvement is remarkable in that \newcite{irtg-ctf-17} found that
coarse-to-fine parsing, which also prunes the substrings a
finer-grained parser considers, did not improve TAG parsing
performance.

\setlength{\tabcolsep}{4pt}

\begin{figure}
  \centering
  {\small
  \begin{tabular}{cl|rrrr}
    \hline
    &Parser  & f-score & time & speedup & \% gold\\\hline
    \parbox[t]{2mm}{\multirow{4}{*}{\rotatebox[origin=c]{90}{binarized}}}
    & Unpruned & 51.4 & 9483 & 1.0x & 5.3\\ 
    & CC ($\theta=0.95$) & 53.6 & 2489 & 3.8x & 76.7\\ 
    & CC (gold) & 53.9 & 2281 & 4.2x & 100.0\\ 
    & supertag ($k=3$) & 77.5 & 137 & 69.4x & 29.7\\ 
    \hline
    \parbox[t]{2mm}{\multirow{12}{*}{\rotatebox[origin=c]{90}{unbinarized}}} 
    & supertag ($k=3$) & 78.5 & 132 & 72.0x & 30.2\\ 
    & \ldots\ + CC (0.95) & 78.4 & 76 & \textbf{124.3x} & 91.6\\ 
    & \ldots\ + CC (0.99) & 79.2 & 80 & 119.2x & 86.1\\ 
    & \ldots\ + CC (gold) & 78.3 & 74 & 127.9x & 100.0\\ 
    \cdashline{2-6}[1pt/2pt]
    & \ldots\ + B/E (0.95) & 79.2 & 87 & 108.9x & 74.5\\ 
    & \ldots\ + B/E (0.8) & 78.4 & 84 & 113.3x & 76.9\\ 
    \cline{2-6}
    & supertag ($k=10$) & 79.4 & 1768 & 5.4x & 1.5\\ 
    & \ldots\ + CC (0.95) & 80.6 & 265 & 35.8x & 71.3\\ 
    & \ldots\ + CC (0.99) & 81.0 & 288 & 33.0x & 60.3\\ 
    & \ldots\ + CC (gold) & 81.9 & 252 & 37.6x & 100.0\\ 
    \cdashline{2-6}[1pt/2pt]
    & \ldots\ + B/E (0.95) & 81.1 & 397 & 23.9x & 35.6\\ 
    & \ldots\ + B/E (0.8) & 80.7 & 386 & 24.6x & 38.6\\ 
    \hline
%
  \end{tabular}
}
  \caption{Results for TAG parsing.}
  \label{fig:tag-table}
\end{figure}

\paragraph{Supertagging.}
We then investigated the combination of chart constraints with a
neural supertagger along the lines of
\newcite{lewis16:_lstm_ccg_parsin}. We modified the output layer of
Fig.~\ref{fig:adjunction}a such that it predicts the supertag (=
unlexicalized elementary tree) for each token. Each input token is
represented by a 200D GloVe embedding. 

To parse a sentence \w\ of length $n$, we ran the trained supertagger
on \w\ and extracted the top $k$ supertags for each token $w_i$ of
\w. We then ran the Alto PTAG parser on an artificial string
``\textit{1 2 \ldots $n$}'' and a sentence-specific TAG grammar which
contains, for each $i$, the top $k$ elementary trees for $w_i$,
lexicalized with the ``word'' $i$ and weighted with the probability of
its supertag. This allowed us to use the unmodified Alto parser, while
avoiding the possible mixing of supertags for multiple occurrences of
the same word. We then obtained the best parse trees for the original
sentence \w\ by replacing each artificial token $i$ in the parse tree
by the original token $w_i$.

The sentence-specific grammars are so small that we can parse the test
corpus without binarizing them. As Fig.~\ref{fig:tag-table} indicates,
supertagging speeds up the parser by 5x ($k=10$) to 70x ($k=3$); the
use of word embeddings boosts the coverage to almost 100\% and the
f-score to around 80. Adding chart constraints on top of supertagging
further improves the parser, yielding the best speed (at $k=3$) and
accuracy (at $k=10$). We achieve an overall speedup of two orders of
magnitude with a drastic increase in accuracy.

\paragraph{Allowable items for TAG.}
Instead of requiring that a TAG chart item is only allowable if
neither the string $[i,l]$ nor its gap $[j,k]$ violate a chart
constraint (as in Section~\ref{sec:pruning-tag}), one could instead
adopt a simpler definition by which a TAG chart item is allowable if
$i$ and $l$ satisfy the chart constraints, regardless of the
gap.\footnote{We thank an anonymous reviewer for suggesting this
  comparison.}

We evaluated the original definition from
Section~\ref{sec:pruning-tag} (``CC'') against this baseline
definition (``B/E''). As the results in Fig.~\ref{fig:tag-table}
indicate, the B/E strategy achieves higher accuracy and lower parsing
speeds than the CC strategy at equal values of $\theta$. This is to be
expected, because CC has more opportunities to prune chart items
early, but false positive chart constraints can cause it to
overprune. When $\theta$ is scaled so both strategies achieve the same
accuracy -- i.e., B/E $\theta=0.8$ for CC $\theta=0.95$, or CC
$\theta=0.99$ for B/E $\theta=0.95$ --, CC is faster than B/E. This
suggests that imposing chart constraints on the gap is beneficial and
illustrates the flexibility and power of the ``admissible items''
approach we introduce here.


\subsection{Discussion}

The effect of using chart constraints is that the parser considers
fewer substructures of the input object -- potentially to the point
that the asymptotic parsing complexity is reduced below that of the
underlying grammar formalism
\cite{roark12:_finit_state_chart_const_reduc}. In practice, we observe
that the percentage of chart items whose begin positions and end
positions are consistent with the gold standard tree (``\% gold'' in
the figures) is increased by CTF and supertagging, indicating that
these suppress
the computation of many spans that
are not needed for the best tree. However, chart constraints prune
useless spans out much more directly and completely, leading to a
further boost in parsing speed.

\begin{figure}
  \centering
  \includegraphics[width=\columnwidth]{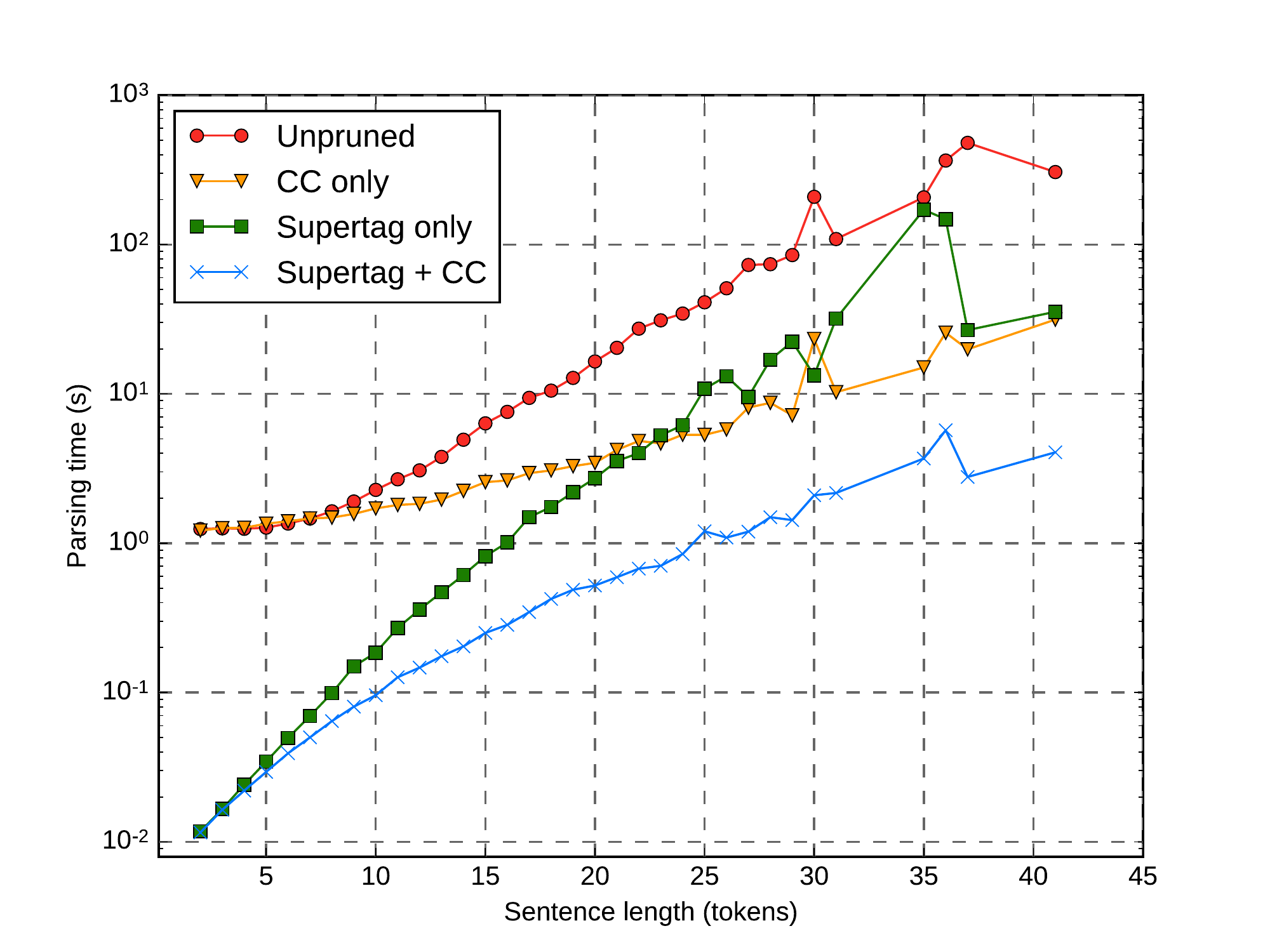}
  \caption{TAG parsing speed as a function of sentence length.}
  \label{fig:lengths-times}
\end{figure}

Because we remove multiple adjunctions in the TAG experiment, most
sentences in the corpus are shorter than in the original. This might
skew the parsing results in favor of pruning techniques that work best
on short sentences. We checked this by plotting sentence lengths
against mean parsing times for a number of pruning methods in
Fig.~\ref{fig:lengths-times} (supertagging with $k = 10$, chart
  constraints with $\theta = 0.95$). As the sentence length increases,
parsing times of supertagging together with chart constraints grows
much more slowly than the other methods. Thus we can expect the
relative speedup to increase for corpora of longer sentences.

\section{Conclusion} \label{sec:conclusion}

Chart constraints, computed by a neural tagger, robustly accelerate
parsers both for PCFGs and for more expressive formalisms such as
TAG. Even highly effective pruning techniques such as CTF and
supertagging can be further improved through chart constraints,
indicating that they target different sources of complexity.


By interpreting chart constraints in terms of allowable chart items,
we can apply them to arbitrary chart parsers, including ones for
grammar formalisms that describe objects other than strings, e.g.\
graphs \cite{ChiangABHJK13,GroschwitzKT15}. The primary challenge here
is to develop a high-precision tagger that identifies allowable
subgraphs, which requires moving beyond LSTMs.


An intriguing question is to what extent chart constraints can speed
up parsing algorithms that do not use charts. It is known that chart
constraints can speed up context-free shift-reduce parsers
\cite{COIN:COIN12094}. It would be interesting to see how a neural
parser, such as \cite{DyerKBS16}, would benefit from chart constraints
calculated by a neural tagger.


\paragraph{Acknowledgments.} We are grateful to Jonas Groschwitz,
Christoph Teichmann, Stefan Thater, and the anonymous reviewers for
discussions and comments. This work was supported through the DFG
grant KO 2916/2-1.



\clearpage
\appendix
\section{Training details}

Both neural networks were implemented using Tensorflow 1.1.0.

\subsection{Chart constraints}
The network has two hidden layers consisting of Tensorflow LSTM cells
with 100 units each. Weights are initialized by sampling from a
uniform probability distribution with values between $-0.1$ and
$0.1$. No dropout is applied between layers.

As input, the network uses 100-dimensional pre-trained word embeddings
and a one-hot encoding for POS tags. Word embeddings for unknown words
(UNK) and numbers (NUMBER) were initialized using a random normal
distribution with a standard deviation of 0.5. Input sentences are
processed one-by-one, i.e. no batching is performed.

We used the RMSProp optimizer for training, with a starting learning
rate of $5 \cdot 10^{-4}$. The learning rate was decreased by 10\%
after each training epoch. The training process was stopped after 6
epochs, when accuracy on the development set stopped increasing. On an
AMD Opteron 6380 processor with a clock rate of 2.5 GHz, the training
process took about 4 hours in total. Tagging the entire test set takes
about 10 seconds.

\subsection{Supertagging}
The network has two hidden layers consisting of Tensorflow LSTM
cells. The first layer consists of 200 units, the second layer of 100
units. Weights are initialized by sampling from a uniform probability
distribution with values between $-0.1$ and $0.1$. A dropout of 50\%
is applied between layers during training.

As input, the network uses 200-dimensional pre-trained word
embeddings. Word embeddings for unknown words (UNK) and numbers
(NUMBER) were initialized using a random normal distribution with a
standard deviation of 0.5. The network does not use POS tags as
input. Input sentences are processed one-by-one, i.e. no batching is
performed.

We used the Adam optimizer for training, with a starting learning rate
of $5 \cdot 10^{-4}$. The learning rate was decreased by 10\% after
each training epoch. The training process was stopped after 6 epochs,
when accuracy on the development set stopped increasing. On an AMD
Opteron 6380 processor with a clock rate of 2.5 GHz, the training
process took about 11 hours in total. Tagging the entire test set
takes about 10 seconds.


\bibliographystyle{acl}
\bibliography{mybib}


\end{document}